# Self-Inspection Method of Unmanned Aerial Vehicles in Power Plants Using Deep Q-Network Reinforcement Learning

*A Deep-Q network application for autonomous path planning*

Haoran Guan
Department of Mechanical and Industrial Engineering
University of Toronto
Toronto, Canada

haoran.guan@mail.utoronto.ca

* Corresponding author: haoran.guan@mail.utoronto.ca

**Abstract** – For the purpose of inspecting power plants, autonomous robots can be built using reinforcement learning techniques. The method replicates the environment and employs a simple reinforcement learning (RL) algorithm. This strategy might be applied in several sectors, including the electricity generation sector. A pre-trained model with perception, planning, and action is suggested by the research. To address optimization problems, such as the Unmanned Aerial Vehicle (UAV) navigation problem, Deep Q-network (DQN), a reinforcement learning-based framework that Deepmind launched in 2015, incorporates both deep learning and Q-learning. To overcome problems with current procedures, the research proposes a power plant inspection system incorporating UAV autonomous navigation and DQN reinforcement learning. These training processes set reward functions with reference to states and consider both internal and external effect factors, which distinguishes them from other reinforcement learning training techniques now in use. The key components of the reinforcement learning segment of the technique, for instance, introduce states such as the simulation of a wind field, the battery charge level of an unmanned aerial vehicle, the height the UAV reached, etc. The trained model makes it more likely that the inspection strategy will be applied in practice by enabling the UAV to move around on its own in difficult environments. The average score of the model converges to 9,000. The trained model allowed the UAV to make the fewest number of rotations necessary to go to the target point.

## I. Introduction

This study will show an innovative application for unmanned aerial vehicle (UAV) autonomous navigation employing a reinforcement learning (RL) trained model in a simulated power plant environment, with realistic element settings such as battery charge level, wind field, etc. under the Deep-Q Network (DQN) framework. The trained model increases the likelihood that the inspection approach will be used in practice by allowing the UAV to travel independently in challenging settings.

Power plant inspection is an occupation with excellent potential for automation through artificial intelligence. The exorbitant expense of putting up cameras and the demand for ongoing maintenance has led to an increase in the use of UAV for self-inspection. [1] Additionally, it is perilous to have technicians in the area (circulating) because mishaps still happen in manufacturing facilities, requiring specialized training to be swift and exact. Motion planning, object recognition, and path planning are necessary for this traditional UAV navigation. [2] Traditional navigation techniques like artificial potential fields [3] and SAR system-based UAV navigation [4] have challenges when it comes to simulating the environment. These challenges can be solved by using RL algorithms. Additionally, traditional navigation methods are less flexible and cost more to apply with sensors.

Reinforcement learning, a subfield of machine learning, has garnered a lot of interest in both academia and business. The reinforcement learning strategy, could also be less computationally demanding than the traditional path planning method. In a way, a pre-trained model with perception, planning, and action will enable the UAV to quickly discover the optimal route to the target.

Numerous studies on reinforcement learning algorithms have been conducted in various disciplines, including UAV applications. [5] [6] Jonathan M. Aitken et al. [7] suggest a way of using autonomous robots to inspect water and sewer pipe networks with localization and mapping for the replacement of manual inspection. Xiaoping Jia et al. [8] propose an application approach of intelligent control technology in the inspection robot in a thermal power plant. Huy X. Pham et al. [9] conducted a simulation and implementation of the UAV system using reinforcement learning.

Deep Q-network [10], a reinforcement learning-based framework introduced by Deepmind in 2015, combines both deep learning and Q-learning and is a potential approach to tackle optimization issues, such as the UAV navigation problem. The model needs to employ a Q function, which assesses the predicted discounted benefit for choosing a certain action at a given state, rather than a Q table in the proposed UAV performance scenario.

The research suggests a power plant inspection system with UAV autonomous navigation using DQN reinforcement learning to address the issues with existing practices. These training procedures differ from existing reinforcement learning training methods in that they set reward functions with reference to states and consider both internal and exterior effect elements. For instance, the essential parts of the reinforcement learning section of the approach introduce states like the simulation of a wind field, the charge level of an unmanned aerial vehicle's battery, the height the UAV reached, etc. The trained model increases the likelihood that the inspection approach will be used in practice by allowing the UAV to travel independently in challenging settings.

## II. METHOD

### A. UAV Self-Inspection Model

The navigation of the UAV during the inspection is the primary subject of this paper. The power plant, which contains buildings and wires, is where the inspection is conducted. [Figure 1] The model is used to depict the power plant as well as the internal and exterior environmental conditions, including the amount of wind and charge level, among others. In order to update statuses and rewards, the UAV itself and the environment it is in both contribute to perception. [Figure 2]

Figure 1. UAV navigation in power plant environment with reinforcement learning model

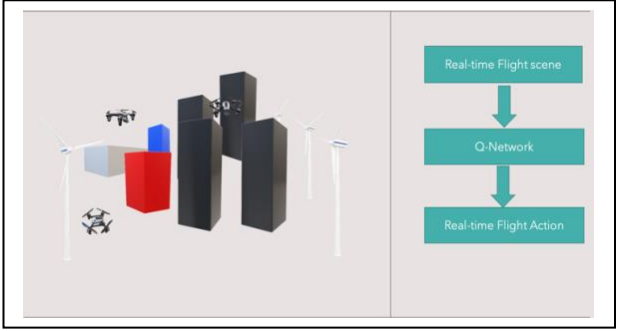

Figure 2. Perception model of the UAV

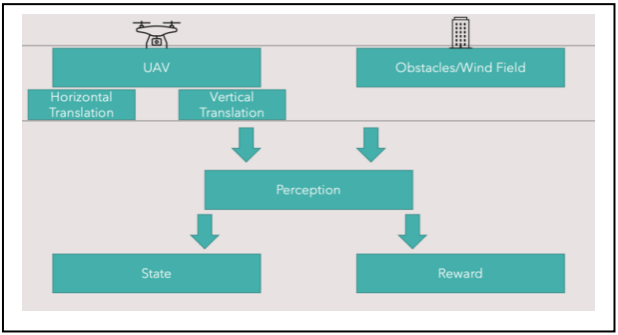

### B. Key Elements of Deep-Q Network Reinforcement Learning

Markov decision processes may be specifically described as reinforcement learning (MDPs). They have four components: a policy, a reward signal, an environment, and a utility function. This makes it a strong contender for handling complicated problems and capturing real-world circumstances. This element will focus on the utility function which is the Q function detail.

However, traditional reinforcement learning calls for the agents to produce prior knowledge for the new state and adopt the proper representations of the environment based on high-dimensional input. Its usefulness is limited to the low-dimensional space where the characteristics may be used to their full potential. Deep reinforcement learning has developed into a high performing tool for solving complicated issues, and DQN, which incorporates deep neural networks into reinforcement learning, was created to fill in these gaps. [10] Moreover, DQN is a model-free, value-based, off-policy method whenever using the established strategies and updating the Q value. State space and action spaces are limited sizes to decrease computational power. $Q(s, a)$ will be represented by a function, not a table. Next is the pseudocode for the DQN method that demonstrates details. [Table 1]

TABLE I. DEEP Q-LEARNING NETWORK

| *Algorithm 1: Deep Q-learning WITH EXPERIENCE REPLAY* |
|---|
| Initialize replay memory $D$ to capacity $N$ |
| Initialize action-value function $Q$ with random weights $\theta$ |
| Initialize target action-value function $\mathbb{Q}$ with weights $\dot{\theta} = \theta$ |
| **for** episode 1, M do initialize sequence $s_1 = \{x_1\}$ and preprocessed sequence $\varphi_1 = \varphi(s_1)$ |
|     **for** t = 1. T do |
|         Boolean step \| **if** with probability $\varepsilon$ select a random action $a_t$ |
|         Otherwise, select $a_t = arg\ max_a Q(\varphi(s_t), a; \theta)$ |
|         Execute $a_t$ in emulator and observe reward $r_t$ and image $x_{t+1}$ |
|         Let $s_{t+1} = s_t, a_t, x_{t+1}$ and preprocess $\varphi_{t+1} = \varphi(s_{t+1})$ |
|         Store experience $\varphi_t, a_t, r_t, \varphi_{t+1}$ in D |
|         Sample random minibatch of experiences $\varphi_j, a_j, r_j, \varphi_{j+1}$ from D |
|         Boolean Step \| **if** episode terminates at step $j + 1$, $y_j = r_j$ |
|         Otherwise, perform calculation $y_j = r_j + \gamma max_a, \mathbb{Q}(\varphi_{j+1}, a'; \dot{\theta})$ |
|         Perform a gradient descent step on $(y_j = Q(\varphi_j, a_j; \theta))^2$ with respect to weights $\theta$ |
|         Reset $\mathbb{Q} = Q$ |
|     End **for** t loop |
| End **for** episode loop |

The UAV agent is trained using the curriculum learning approach in the three-dimensional space of x100 y100 z22, and the gradient training of the agent is done using preset courses of varying degrees of difficulty so that the agent can gain decision-making expertise more quickly. It is required to randomly select behaviors to assess the environment since early training lacks decision-making experience. The random testing time in this study is set to 4000, and agent actions are chosen using the greedy approach. The period's greedy probability gradually drops from 1 to 0.01. 4000 episodes later the likelihood of being greedy is still 0.01.

*C. Environment*

The environment is created using Matplotlib and Axes 3D library functions in Python. This major body construction consists of three steps that produce impediments to path planning. There are three levels of considerations in play in this section. The first is the size of the planning space; it is not feasible to design a highly decentralized space due to both practical and computational reasons. A cubic area is defined as having a length of 100km, a height of 100km, and a width of 22km to condense the modeling space. The second step is to establish a wind field with a wind speed that is initially set at 30 km/h. This modeling can be further complicated instead of setting a fixed number, specifying a function that causes the wind speed to change over time and increase with height. The last step in defining the construction pieces is to specify their render color, size, and center coordinates. [Figure 3]

Figure 3. Simulation space (unit: km)

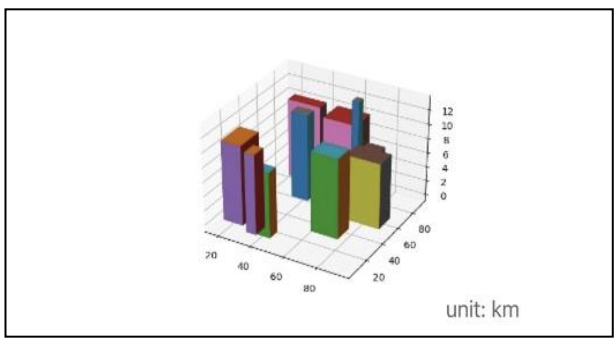

*D. Reward*

The drone's total reward may be computed by adding the climb reward and target reward. However, some additional reward and punishments are provided at various termination stages, such as terminating at the goal point with a collision or having the battery run out without a collision, etc. These termination stages are considered with each respective reward modification. Here, avoiding collisions is of utmost importance to reach the targeted spot. As a result, the total reward will be raised or lowered to a clear stage depending on the reward adjustment for achieving the goal location and a collision incidence. However, as none of these events will have a significant impact on the ultimate termination stage, situations like going over the maximum number of steps or running out of battery are given a tiny reward modification. [Table 2]

TABLE II. REWARD CALCULATION ALGORITHM

| *Algorithm 2: Reward calculation in different end stages* |
|---|
| Initialize clime reward $r_{climb} = wc \times (|self.z - self.target[2]|)$ |
| Initialize target reward $r_{target}$ with respect to distance |
| Total reward $r_{total} = r_{climb} + r_{target}$ |
| Boolean step \| **if** the UAV did not get to the target NOR crash NOR exceed the maximum steps it can take in the space |
|     Calculate total reward |
|     if there is a crash and collision, return $r_{total} - 500$ |
|     if the UAV gets to the target position, return $r_{total} + 500$ |
|     if the UAV exceeds the maximum steps it can take in space, return $r_{total} - 30$ |
|     If the battery run out, return $r_{total} - 30$ |

*E. States and Actions*

The state is defined by initializing the various state parameters of the drone and updating with calculation of its motion change. States play a significant role in determining the reward. One of the reward conditions, for instance, must specify that if the UAV reaches the intended place. The coordinates position and target coordinates must be recorded in the states and the UAV coordinate position must be updated concurrently with the motion.

The procedure updates the state by calculating the motion change from the action value. The change value might be computed by using the update algorithm's functions. Another penalty may be applied if the UAV does not always move, and the change values are equal to zero. Other states, such as the UAV's energy consumption, the distance to obstacles, the steps that the UAV has made, etc., may be updated using the amended UAV coordinates.

The indicators for fundamental energy consumption, current energy use, and spent energy are all included in the state. For calculating rewards, other states like range detected, surrounding barriers, crash probability, distance steps taken, goal distance, etc. are also collected.

Prior to utilizing actions to update the states, action functions $a_t$ are probabilistically chosen at random $\varepsilon$ using the greedy strategy. After training, that greedy approach drops from 1 and converges to 0.01. This is done to address the issue that the model cannot choose the best course of action when the overall scores have very significant negative values.

## III. RESULTS

After 40,000 episodes of training on top of a pretrained model, the sum of episodes trained are 173,399 episodes. The model reaches a steady average score around 9,000 proved by the average score plot in the last 5000 episodes of training. [Figure 3] The training ends at 40000 since the average score has been declining for the previous two thousand training sessions rather than increasing. [Figure 4]

With the aid of a trained model and a randomly generated environment, the UAV successfully took a direct path to the target location with the fewest number of turns feasible.

Figure 4. Average score plot for the last 5000 episodes

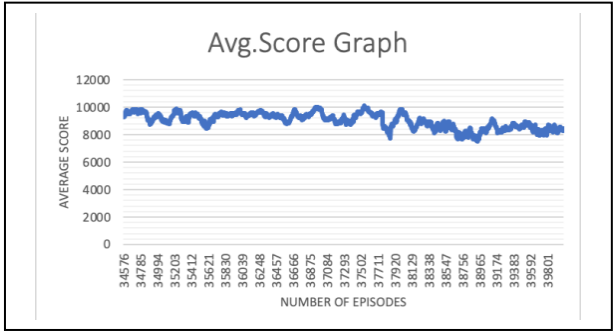

After 360 episodes of training, the graph's score considerably increased and converged. The average score plot displays a smooth curve with an upward trend near 10,000. [Figure 5]

Figure 5. Average score plot for the first 360 episodes

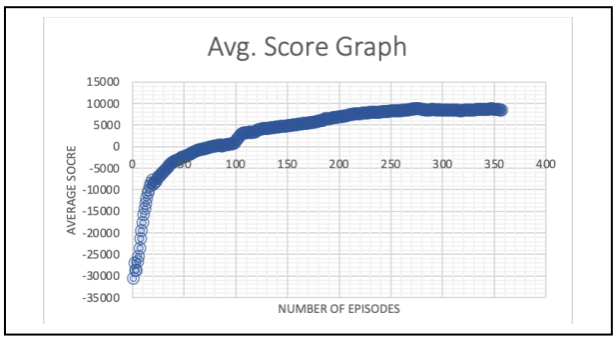

A comparison was made using the pre-trained model and the model that is been trained after 173,399 episodes. [Figure 6] and [Figure 7] As can be seen in the graph, the pre-trained model still causes the UAV to crash into the building and fly through to the desired location, but the trained model performs significantly better with time.

Figure 6. A path navigated using the pre-trained model in simulated space

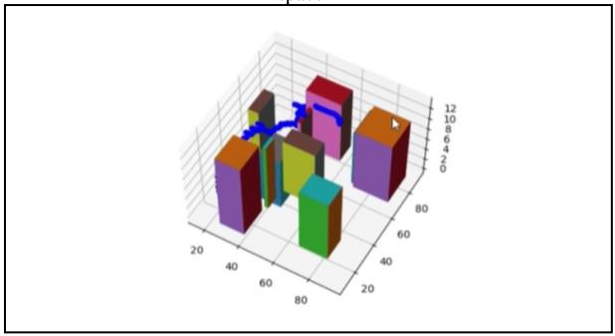

Figure 7. A path navigated using the converged model in simulated space

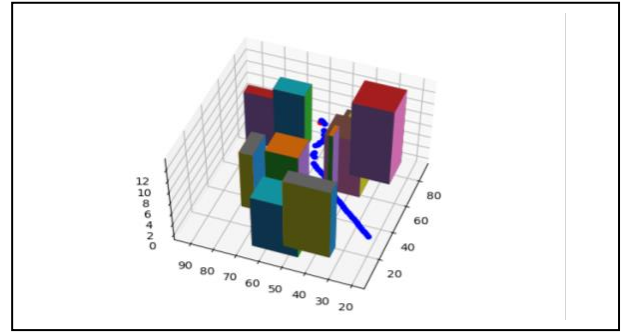

IV. DISCUSSION

The discussion will address the reasons behind why, occasionally, the score plummets to a severely negative number at the closing of the training sessions. [Figure 8] The average score alone cannot serve as an accurate indicator of achievement. It is merely a metric for determining whether to continue training the model. The risk of a building crash cannot be tolerated by UAV navigation since it would result in a significant financial loss.

Figure 8. Score graph for the last 5000 episodes

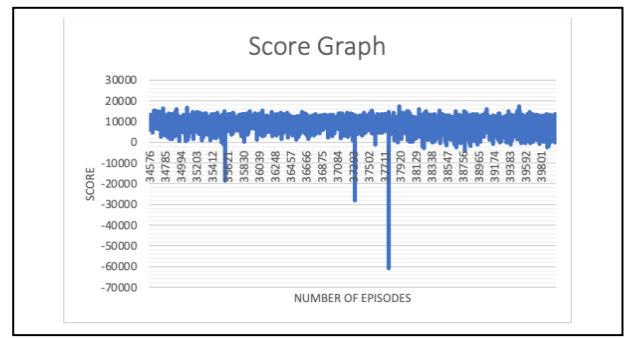

The fact that the target generation positions might occasionally be quite near to the building and the drone has no choice but to crash into the building to reach the target place is one possible source of mistakes. However, both will lead to severe punishments and rewarding outcomes, which is how things balance out. Increasing the penalties for crashing and making it the major goal of navigation, with reaching the target point as the second primary goal, is one possible solution to address this cause of inaccuracy.

The strong wind from below may trigger another source of inaccuracy by pushing the UAV into the building. However, because training takes a very long time, turning off the wings field and then waiting to see how the score will be during the training period is not the best strategy to adopt. After the model has been created for at least 1,000 episodes, it is feasible to test this strategy by disabling the wind field in real-time flight navigation instances.

## V. Conclusion

The strategy described in the study uses the DQN reinforcement learning method to teach a UAV to learn to navigate to the intended position randomly generated in a model of a power plant that includes a wind field and considers energy usage. The simulation of the outcome demonstrates that this method can offer a template for a UAV to navigate to the intended place while maintaining awareness of itself and its surroundings. The training score plot which converges at 10,000 provides evidence about the experiment's methodology. Additionally, this article makes some suggestions for prospective upgrades to real-world deployment. For instance, provide a feasible verification mechanism to explain for outlying scores or develop a wind field function to match the external fluid effect seen in the actual world to prevent mistakes.